\documentclass[10pt,twocolumn,letterpaper]{article}

\usepackage{cvpr}
\usepackage{times}
\usepackage{epsfig}
\usepackage{graphicx}
\usepackage{epstopdf}\usepackage{amsmath}
\usepackage{amssymb}

\cvprfinalcopy 

\def\cvprPaperID{6518} 

\ifcvprfinal\fi
\begin{document}

\title{Adversarial Defense of Image Classification Using a Variational Auto-Encoder}

\author{Yi Luo\\
Duke University\\
Durham, NC, USA\\
{\tt\small yi.luo4@duke.edu}
\and
Henry Pfister\\
Duke University\\
Durham, NC, USA\\
{\tt\small henry.pfister@duke.edu}
}

\maketitle
\begin{figure*}[h]
  \centering
  \begin{tabular}{@{}ccc@{}}
    \includegraphics[width=.3\linewidth]{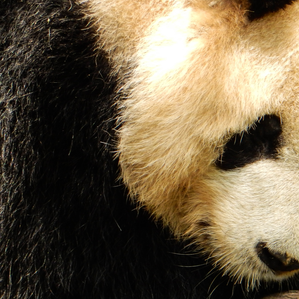}  & \includegraphics[width=.3\linewidth]{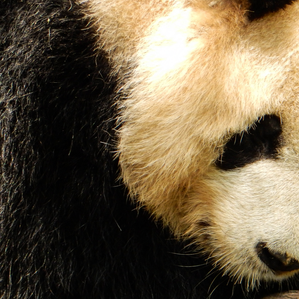}  &
    \includegraphics[width=.3\linewidth]{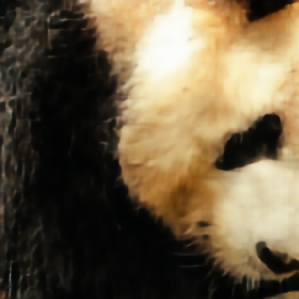}\\[\abovecaptionskip]
    \small (a) "Panda" Original image. & \small (b) "Dog" Adversarial image. & \small (c) "Panda" VAE reconstructed image.
  \end{tabular}

  \vspace{\floatsep}
  \caption{Classification results of original image, adversarial image and VAE reconstructed image.}\label{fig:intro}
\end{figure*}

\begin{abstract}
   Deep neural networks are known to be vulnerable to adversarial attacks.
   This exposes them to potential exploits in security-sensitive applications and highlights their lack of robustness. This paper uses a variational auto-encoder (VAE) to defend against adversarial attacks for image classification tasks. This VAE defense has a few nice properties: (1) it is quite flexible and its use of randomness makes it harder to attack; (2) it can learn disentangled representations that prevent blurry reconstruction; and (3) a patch-wise VAE defense strategy is used that does not require retraining for different size images. For moderate to severe attacks, this system outperforms or closely matches the performance of JPEG compression, with the best quality parameter.  It also has more flexibility and potential for improvement via training.
  
\end{abstract}


\section{Introduction}
Deep neural network (DNNs) now provide state-of-the-art performance in many tasks related to computer vision and pattern recognition \cite{goodfellow2016deep}. As the applications of deep learning expand, DNNs are more likely to be used in security-sensitive systems, such as medical imaging, autonomous driving, and surveillance systems. The reliability and robustness of deep learning for these applications is essential and cannot be ignored. A DNN, like many other systems, can be attacked purposefully using carefully crafted methods. For example, results have shown that DNNs are vulnerable against a variety of \textit{adversarial attacks} \cite{carlini2017towards, fawzi2018analysis, goodfellow6572explaining, nguyen2015deep,szegedy2013intriguing}. 

Adversarial attacks are designed to maliciously add small perturbations to the original input in order to fool the neural network into making incorrect predictions. For typical adversarial attacks, a key element is that the adversary knows in advance which classifier will be used and can design the perturbation using knowledge of the classifier.
Of course, there is a trade-off between the size of the perturbation and its effectiveness.
In many cases, the adversary can greatly increase the classification error rate by adding a perturbation that is almost imperceptible~\cite{carlini2017adversarial,liang2018detecting}.

In an abstract sense, one can model the adversarial classification problem as a two-player zero-sum game. In this game, the attacking player wins if the classifier makes an error and the defending player wins if the classifier is correct.
A single point is awarded to the winner and no points are awarded to the loser.
It is worth noting that the attacker has an advantage because their move depends on the true image while the defender's move can only depend on the perturbed image.
This is a game with \emph{imperfect information} because the defender does not see the original image.
In games of this type, it is well-known that both players may benefit from using randomized strategies.
This is because, if one player fixes their strategy, then the other can always optimize against it.
Although current adversarial models typically consider less complicated scenarios, this framework motivated us to choose a VAE due to its use of randomness.
The asymmetry between the attack and defense in current adversarial models is also apparent in paper that design defense mechanisms for varied attacks~\cite{huang2017safety,zantedeschi2017efficient}.

Adversarial training methods, which retrain the original neural network with additional adversarial examples~\cite{goodfellow6572explaining}, can learn to defend against specific attacks on which they are trained.
But, as mentioned earlier, attackers still have an advantage because the cost of altering the attack is much lower than retraining with new adversarial examples.
Therefore, researchers continue to search for a universal defense that performs well against a wide range of attacks.
One example of an efficient low-cost defense is JPEG compression~\cite{das2017keeping}.

Meanwhile, recent works have also shown that randomness can be quite effective for defense~\cite{das2018shield, zhou2018breaking}. The contributions of our paper can be listed as follows:
\begin{itemize}
   \item A novel defense is proposed for adversarial attacks on image classification networks which uses a variational auto-encoder (VAE) to reconstruct the input image before classification by the targeted neural network model.
   This defense method does not modify the deployed network nor does it depend on the particular attack chosen.  Thus, it is universal.
   \item Our defense strategy uses randomness in our via the random sampling process in the VAE.
   Patch-wise defense is used for large images to reducing the training cost of the VAE.
   \item The proposed method is flexible and has multiple tunable hyper-parameters.
   Even without retraining the VAE, the defense can be altered by modifying the reconstruction process.
   Experiments show that this defense is capable of matching the performance other defenses based on JPEG compression.
   Due to its flexibility, however, it has more potential for improvement and integration with other methods.
\end{itemize}

\section{Related Work}
\subsection{Adversarial Attacks}

Consider a classification problem for images $x\in\mathbb{R}^n$ where there are $m$ classes labeled by $[m]\triangleq \{1,2,\ldots,m\}$.
For this problem, a neural network $f(x;\theta)$ with parameters $\theta \in \Theta$ maps an input image $x\in \mathbb{R}^n$ to a vector $z\in \mathbb{R}^m$ and the index $\hat{y}$ of the largest value in $z$ is returned as the estimated class label.
Training is typically handled by using some form of stochastic gradient descent (SGD) to iteratively update $\theta$ in a way that reduces the training loss
\[ \mathcal{L}(\theta) = \sum_{(x,y)\in \mathcal{D}} \ell (f(x;\theta),e_y), \]
where $\mathcal{D}$ is the set of training pairs, $e_y \in \mathbb{R}^m$ denotes the standard basis vector with a one in the $y$-th position, and $\ell(z,\hat{z})$ is the loss function associated with the neural network outputting $z$ when the true one-hot class vector is $\hat{z}$.
Common choices are squared-error $\ell(z,\hat{z})=\sum_{i=1}^m |z_i-\hat{z}_i|^2$ and cross-entropy $\ell(z,\hat{z})=-\sum_{i=1}^m \hat{z}_i \log z_i$.

For a particular input $x$ with true class $y$, an adversarial attack on this network adds a perturbation to create $\hat{x}$ such that $\| x -\hat{x} \|$ is small and  $\hat{y} \neq y$, where $\| \cdot \|$ is some norm on $\mathbb{R}^n$.
A \emph{targeted} attack chooses the value of $\hat{y}$ in advance while a \emph{non-targeted} attack is free to choose any $\hat{y} \neq y$.

For an input $x$ with class label $y$, many attack methods are based computing the gradient $\nabla_x \ell (f(x;\theta),e_y)$ to find perturbation directions that increase the loss function and, thus, cause the network to misclassify $x$.
In this paper, we mainly focus on the following attack methods:

\begin{table}
\begin{tabular}{lll}
\multicolumn{3}{c}{CNN Classifier Structures}                                                               \\ \hline
\multicolumn{1}{l|}{Dataset}   & \multicolumn{1}{l|}{Layer}                         & Shape                 \\ \hline
\multicolumn{1}{l|}{MNIST}     & \multicolumn{1}{l|}{Input}                         & $28\times28\times1$   \\
\multicolumn{1}{l|}{}          & \multicolumn{1}{l|}{Conv+Maxpooling(2,2)}        & $14\times14\times32$  \\
\multicolumn{1}{l|}{}          & \multicolumn{1}{l|}{Conv+Maxpooling(2,2)}        & $7\times7\times64$    \\
\multicolumn{1}{l|}{}          & \multicolumn{1}{l|}{Dense+Dropout(0.4)}          & 1024                  \\
\multicolumn{1}{l|}{}          & \multicolumn{1}{l|}{Dense}                         & 10                    \\ \hline
\multicolumn{1}{l|}{CIFAR-10}  & \multicolumn{1}{l|}{Input}                         & $32\times32\times3$   \\
\multicolumn{1}{l|}{}          & \multicolumn{1}{l|}{(Conv+BatchNorm)$\times$2} & $32\times32\times32$  \\
\multicolumn{1}{l|}{}          & \multicolumn{1}{l|}{Maxpooling(2,2)}               & $16\times16\times32$  \\
\multicolumn{1}{l|}{}          & \multicolumn{1}{l|}{Dropout(0.2)}                  &                       \\
\multicolumn{1}{l|}{}          & \multicolumn{1}{l|}{(Conv+BatchNorm)$\times$2} & $16\times16\times64$  \\
\multicolumn{1}{l|}{}          & \multicolumn{1}{l|}{Maxpooling(2,2)}               & $8\times8\times64$    \\
\multicolumn{1}{l|}{}          & \multicolumn{1}{l|}{Dropout(0.3)}                  &                       \\
\multicolumn{1}{l|}{}          & \multicolumn{1}{l|}{(Conv+BatchNorm)$\times$2} & $8\times8\times128$   \\
\multicolumn{1}{l|}{}          & \multicolumn{1}{l|}{Maxpooling(2,2)}               & $4\times4\times128$   \\
\multicolumn{1}{l|}{}          & \multicolumn{1}{l|}{Dropout(0.4)}                  &                       \\
\multicolumn{1}{l|}{}          & \multicolumn{1}{l|}{Dense}                         & 10                    \\ \hline
\multicolumn{1}{l|}{NIPS 2017} & \multicolumn{1}{l|}{Input}                         & $299\times299\times3$ \\
\multicolumn{1}{l|}{}          & \multicolumn{1}{l|}{Inception V3 net}              & 1001                 
\end{tabular}
\vspace{1mm}
\caption{CNN classifiers structures for experiment. MNIST and CIFAR-10 models are trained from scratch with Stochastic Gradient Descent (SGD). For NIPS 2017 dataset the model is a pre-trained Inception V3 model.}
   \label{tab:CNN}
\vspace{-5mm}   
\end{table}

\paragraph{Fast Gradient Sign Method (FGSM)~\cite{goodfellow6572explaining}:} FGSM is a gradient-based single-step attack method.
It is quite fast and it generates adversarial images by adding or subtracting a fixed amount from each pixel in the image.
For each pixel, the sign of the perturbation is determined by the sign of the gradient of the loss function with respect to the image.
For an image $x$, the adversarial image would be
$$
\hat{x} = \mathrm{clip}(x + \epsilon \cdot \mathrm{sign} \big(\nabla_x \ell ( f(x;\theta),e_y) \big),
$$
where $\mathrm{clip}$ clips a vector to minimum/maximum pixel values and $sign$ computes for the element-wise sign of the corresponding gradient.

\paragraph{Iterative FGSM (I-FGSM) ~\cite{kurakin2016adversarial}:} I-FGSM is based on repeating the FGSM attack $M$ times with $x_0 = x$,  
$$
x_{m+1} = \mathrm{clip}(x_{m} + (\epsilon/M) \cdot \mathrm{sign} \big(\nabla_x \ell ( f(x_{m};\theta),e_y) \big),
$$
and $\hat{x}= x_M$.
During each iteration, $x_{m+1}$ is created by attacking $x_{m}$ with FGSM method and $\epsilon$ reduced by a factor of $M$.
The adversarial input $\hat{x}$ is the output of the final iteration.

\subsection{Defense}
There are two primary methods of defense against adversarial attacks. The first type modifies the structure of the classifier and/or changes the training procedure~\cite{gu2014towards, papernot2016distillation}. The second type does not change the classifier but instead focuses on modifying the input vector to mitigate attacks~\cite{das2018shield,liang2018detecting,prakash2018deflecting,shaham2018defending}. The goal of the second type is to detect and/or remove adversarial perturbations before passing the data to the classifier. In this case, one needs to design a transformation $T$ that maps images to images and minimizes the effect if adversarial perturbations.

For an adversarial image $\hat{x}$ with true class label $y$, the attack implies that $y\neq \hat{y}$, where $\hat{y}$ denotes the class label estimated by the classifier. When the defense is activated, the system instead passes $T(\hat{x})$ to the classifier and this results in the class vector $\tilde{z}=f(T(\hat{x});\theta)$ and the estimated class label $\tilde{y}$.
If the defense is successful, then $\tilde{y} = y$ and the image is classified correctly.

As mentioned by Shaham \etal \cite{shaham2018defending}, one good choice for $T$ is a basis transformation followed by scaling and/or quantization, which includes defenses such as low-pass filtering and JPEG compression.
The success of these methods is attributed to the fact that they alter the image significantly and tend to reduce adversarial perturbations. 

JPEG Compression~\cite{wallace1992jpeg} is a lossy image compression method that first applies the two-dimensional discrete cosine transform to $8\times 8$ image patches. Then, it quantizes the resulting coefficients (using more bits for lower frequencies) and uses lossless compression to compress the quantized values.
We note that lossless compression plays no role in the adversarial defense and can be ignored in this application.
There is a quality parameter that controls the amount of information loss during the compression. In our experiments, different quality parameters provide different defense performance depending on the attack strength.

\begin{table}[!htbp]
\begin{tabular}{lll}
\multicolumn{3}{c}{Variational Auto-encoder Structures}                                            \\ \hline
\multicolumn{1}{l|}{Dataset}  & \multicolumn{1}{l|}{Layer}                  & Shape                \\ \hline
\multicolumn{1}{l|}{MNIST}    & \multicolumn{1}{l|}{Input}                  & $28\times28\times1$  \\
\multicolumn{1}{l|}{}         & \multicolumn{1}{l|}{Conv+Maxpooling(2,2)} & $14\times14\times16$ \\
\multicolumn{1}{l|}{}         & \multicolumn{1}{l|}{Conv+Maxpooling(2,2)} & $7\times7\times8$    \\
\multicolumn{1}{l|}{}         & \multicolumn{1}{l|}{Dense}                  & 256                  \\ \hline
\multicolumn{1}{l|}{CIFAR-10} & \multicolumn{1}{l|}{Input}                  & $32\times32\times3$  \\
\multicolumn{1}{l|}{}         & \multicolumn{1}{l|}{Conv+Maxpooling(2,2)} & $16\times16\times64$ \\
\multicolumn{1}{l|}{}         & \multicolumn{1}{l|}{Conv}                   & $16\times16\times32$ \\
\multicolumn{1}{l|}{}         & \multicolumn{1}{l|}{Conv}                   & $16\times16\times16$ \\
\multicolumn{1}{l|}{}         & \multicolumn{1}{l|}{Dense}                  & 1024                

  \end{tabular}
  \caption{Variational-autoencoder encoder structures. The decoder structures for each model is symmetric, using transpose convolution and upsampling.}
  \label{tab:VAE}
\end{table}

\begin{table}[!htbp]
\begin{tabular}{lll}
\multicolumn{3}{c}{NIPS 2017 VAE Encoder Structures}                                                      \\ \hline
\multicolumn{1}{l|}{Patch Size}     & \multicolumn{1}{l|}{Layer}                  & Shape                 \\ \hline
\multicolumn{1}{l|}{$16 \times 16$} & \multicolumn{1}{l|}{Input}                  & $16\times16\times3$   \\
\multicolumn{1}{l|}{}               & \multicolumn{1}{l|}{Conv+Maxpooling(2,2)} & $8\times8\times64$    \\
\multicolumn{1}{l|}{}               & \multicolumn{1}{l|}{Conv}                   & $8\times8\times32$    \\
\multicolumn{1}{l|}{}               & \multicolumn{1}{l|}{Conv}                   & $8\times8\times16$    \\
\multicolumn{1}{l|}{}               & \multicolumn{1}{l|}{Dense}                  & 512                   \\ \hline
\multicolumn{1}{l|}{$32\times32$}   & \multicolumn{1}{l|}{Input}                  & $32\times32\times3$   \\
\multicolumn{1}{l|}{}               & \multicolumn{1}{l|}{Conv+Maxpooling(2,2)} & $16\times16\times64$  \\
\multicolumn{1}{l|}{}               & \multicolumn{1}{l|}{Conv}                   & $16\times16\times32$  \\
\multicolumn{1}{l|}{}               & \multicolumn{1}{l|}{Conv}                   & $16\times16\times16$  \\
\multicolumn{1}{l|}{}               & \multicolumn{1}{l|}{Dense}                  & 512                   \\ \hline
\multicolumn{1}{l|}{$64\times64$}   & \multicolumn{1}{l|}{Input}                  & $32\times32\times3$   \\
\multicolumn{1}{l|}{}               & \multicolumn{1}{l|}{Conv+Maxpooling(2,2)} & $32\times32\times128$ \\
\multicolumn{1}{l|}{}               & \multicolumn{1}{l|}{Conv}                   & $32\times32\times64$  \\
\multicolumn{1}{l|}{}               & \multicolumn{1}{l|}{Conv}                   & $32\times32\times32$  \\
\multicolumn{1}{l|}{}               & \multicolumn{1}{l|}{Dense}                  & 1024                 
\end{tabular}
\caption{Variational-autoencoder encoder structures for NIPS 2017 dataset. The decoder structures for each model is symmetric, using transpose convolution and upsampling.}
\label{tab:VAE_NIPS}
\end{table}

\section{Variational Auto-encoder (VAE)}
A variational auto-encoder \cite{kingma2013auto, rezende2014stochastic} is a neural-network model that maps a high-dimensional feature vector to a lower-dimensional latent vector and then incorporates randomness before mapping it back to the original feature space.
It can be seen as a standard auto-encoder, with an encoder and a decoder, but with a random sampling operation separating the two~\cite{baldi2012autoencoders}. 

In our case, we use a VAE where the encoder determines the mean and variance of a Gaussian random vector that is then mapped back to an image by the decoder.
In this case, the encoder is a convolutional neural network parametrized by $\theta$,  denoted as $q_\theta(z|x)$, which maps the input $x\in\mathbb{R}^n$ to the \emph{random} latent vector $z\in\mathbb{R}^d$. The deterministic part of the encoder acts like a compressor~\cite{theis2017lossy} by mapping the key features of the input to a lower-dimensional representation. The decoder is a de-convolutional neural network parametrized by $\phi$, denoted as $p_\phi (x'|z)$ which takes a vector $z$ as input and reconstructs an image $x'\in \mathbb{R}^n$ as the output. Auto-encoders are trained by optimizing a single loss function that measures a distance between input image $x$ and output image $x'$.

The key difference between a VAE and an ordinary auto-encoder is that, instead of the encoder neural network mapping directly to the latent space, it outputs the mean and variance, $\mu_{z}$ and $\sigma_{z}$, of the sampled latent vector $z$.
Then, $z$ is drawn from a Gaussian distribution with mean $\mu_{z}$ and diagonal covariance matrix $\mathrm{diag}(\sigma_{z})$.
Finally, the sampled vector $z$ is passed into the decoder and used to produce the output image $x'$.
Instead of trying to approximate the output image exactly, a VAE defines a conditional distribution that approximates the underlying continuous distribution of images.
Thus, the VAE provides a generative model for images similar to the input image.

In order to learn the parameters of a VAE, the optimized loss function typically consists of two parts: the reconstruction loss and the KL-divergence loss \cite{kullback1997information}.
The first term penalizes the error between the reconstructed image and the input image while the second term encourages the mean/variance pairs for latent variable that are far from a standard Gaussian vector.
The loss function for the $i$-th image in the training set is given by
\begin{align*}
     l_i(\theta,\phi) = &E_{z\sim q_\theta(z|x_i)}[-\log  (p_\phi(x_i|z)]\\
   & + D_{KL}(q_\theta(z|x_i)||p(z)),
\end{align*}
where $p(z)$ is a standard Gaussian distribution on $\mathbb{R}^d$, $q_{\theta} (z|x_i)$ is a Gaussian distribution whose mean and variance are given by the encoder neural network, and $p_\phi (x_i|z)$ is a distribution associated with the reconstruction error term.
For mean-squared error, $p_\phi$ first maps $z$ to $x'$ using the de-convolutional neural network with parameters $\phi$ and then defines
\[ p_\phi (x_i | z) = (2\pi)^{-n/2} \exp (- \|x_i - x'\|^2 / 2),\]
where $\| \cdot \|$ is the Euclidean norm.
Alternatively, one can use binary cross-entropy loss between $x_i$ and $x'$ to define $p_\phi$. 

For the Gaussian case, a "reparameterization trick" allows backpropagation to calculate the gradients during training. The idea is to represent the random sampling process in a differentiable form, where the mean $\mu_z$ is added to standard Gaussian noise which is multiplied by the $\sigma_z$. 

\textbf{Disentanglement}. When we obtain the latent vector compressed by the VAE, we want every component of the latent vector to represent different features of the original image. Disentangled variational auto-encoders are designed to approach this goal \cite{burgess2018understanding,higgins2016beta}. Disentanglement is applied by modifying the optimization loss function and adding the hyperparameter $\beta$ to get
\begin{align*}
     l_i(\theta,\phi) &= -E_{z\sim q_\theta(z|x_i)}[log (p_\phi(x_i|z)]\\
   + &\beta D_{KL}(q_\theta(z|x_i)||p(z))
\end{align*}
In our experiments, we adjusted the $\beta$ hyperparameter to reduce the weight of KL-Divergence loss.
This pushed the VAE towards higher-quality reconstruction with less randomness.
    
\begin{table}[!htbp]

  \centering
  FGSM Attack on MNIST\\[1mm]
  \begin{tabular}{cccc}
 
    \hline
    L2 Diff & No Def & JPEG 23  & Ours\\
    \hline
  
    0.011 & 0.988 & 0.988 & 0.987\\

   0.066 &  0.972 & 0.978  & 0.980\\

0.119 & 0.921   &   0.945 & 0.967 \\

 0.171 & 0.796   &  0.863  & \textbf{ 0.941} \\

 0.220 &  0.573 & 0.686  & \textbf{0.896}\\

0.257 & 0.408   &   0.505  &  \textbf{0.845}\\
    
  \end{tabular}
  \vspace{1mm}
  \caption{Top-1 accuracies on MNIST test set after applying FGSM attack, JPEG defense and our defense. Our defense is very effective compared to JPEG as the perturbation gets larger.}
  \label{tab:MNIST}
  \vspace{-3mm}
\end{table}

\begin{table}[!htbp]
 
  \centering
  FGSM Attack on CIFAR-10\\[1mm]
  \begin{tabular}{cccc}
    \hline
    L2 Diff & No Def & JPEG 23   &  Ours \\
    \hline

0.010  & 0.496   & 0.644 &   0.709\\
 
0.064  & 0.124     &    0.483 &   0.521\\
 
0.118    & 0.105      &   0.323 &  0.347\\

0.171   & 0.098   &   0.217   &   \textbf{0.258}   \\

0.197  & 0.094    &  0.201 &   \textbf{0.233}   \\

  \end{tabular}
  \vspace{1mm}
  \caption{Top-1 accuracies on CIFAR-10 test set. Our VAE defense outperforms JPEG consistently.}
  \label{tab:CIFAR-10}
\vspace{-6mm}
\end{table}

\begin{figure*}[!htbp]
    \centering
    \begin{tabular}{@{}cc@{}}
    \includegraphics[width=0.4\linewidth]{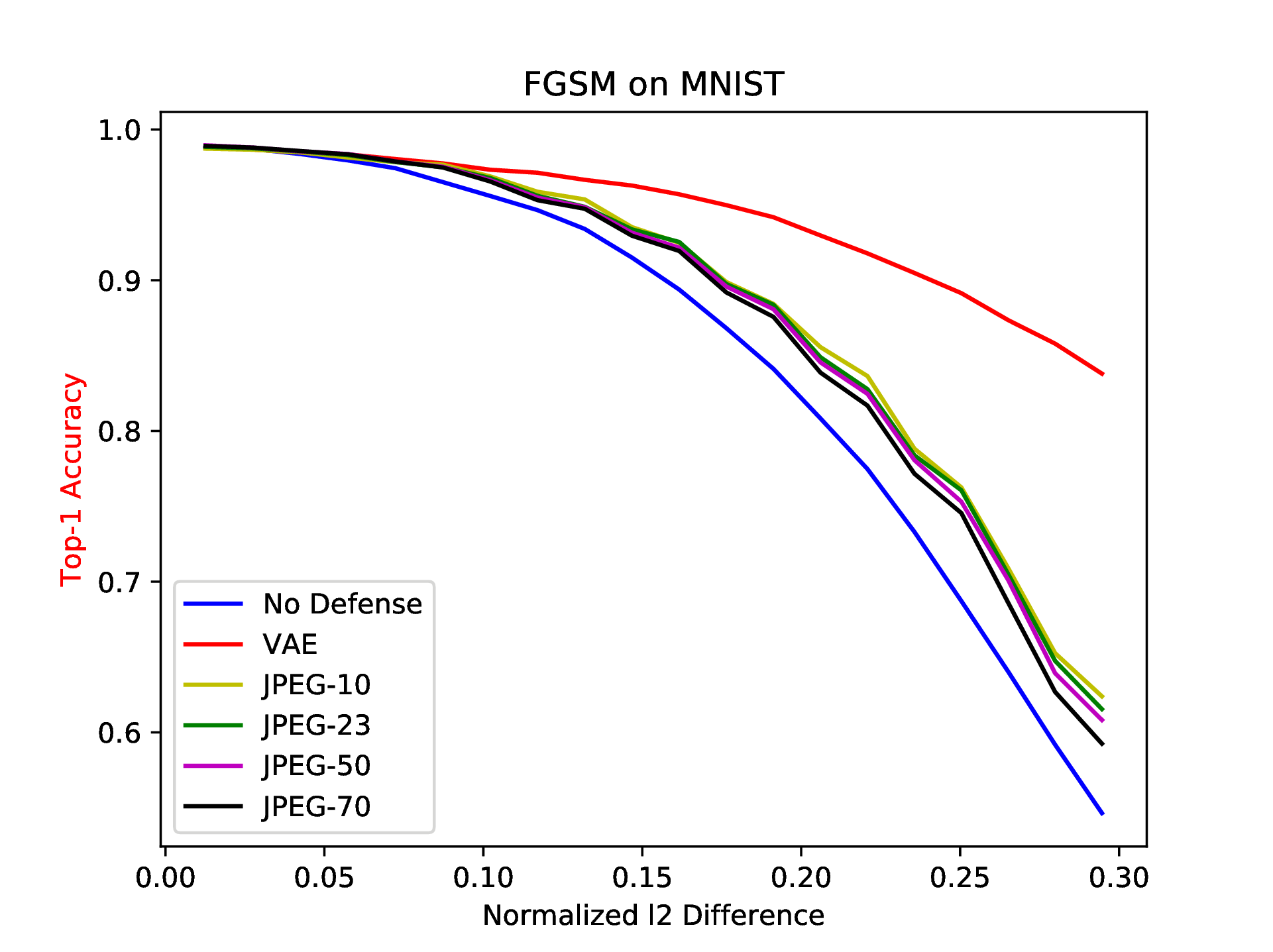} & \includegraphics[width=0.4\linewidth]{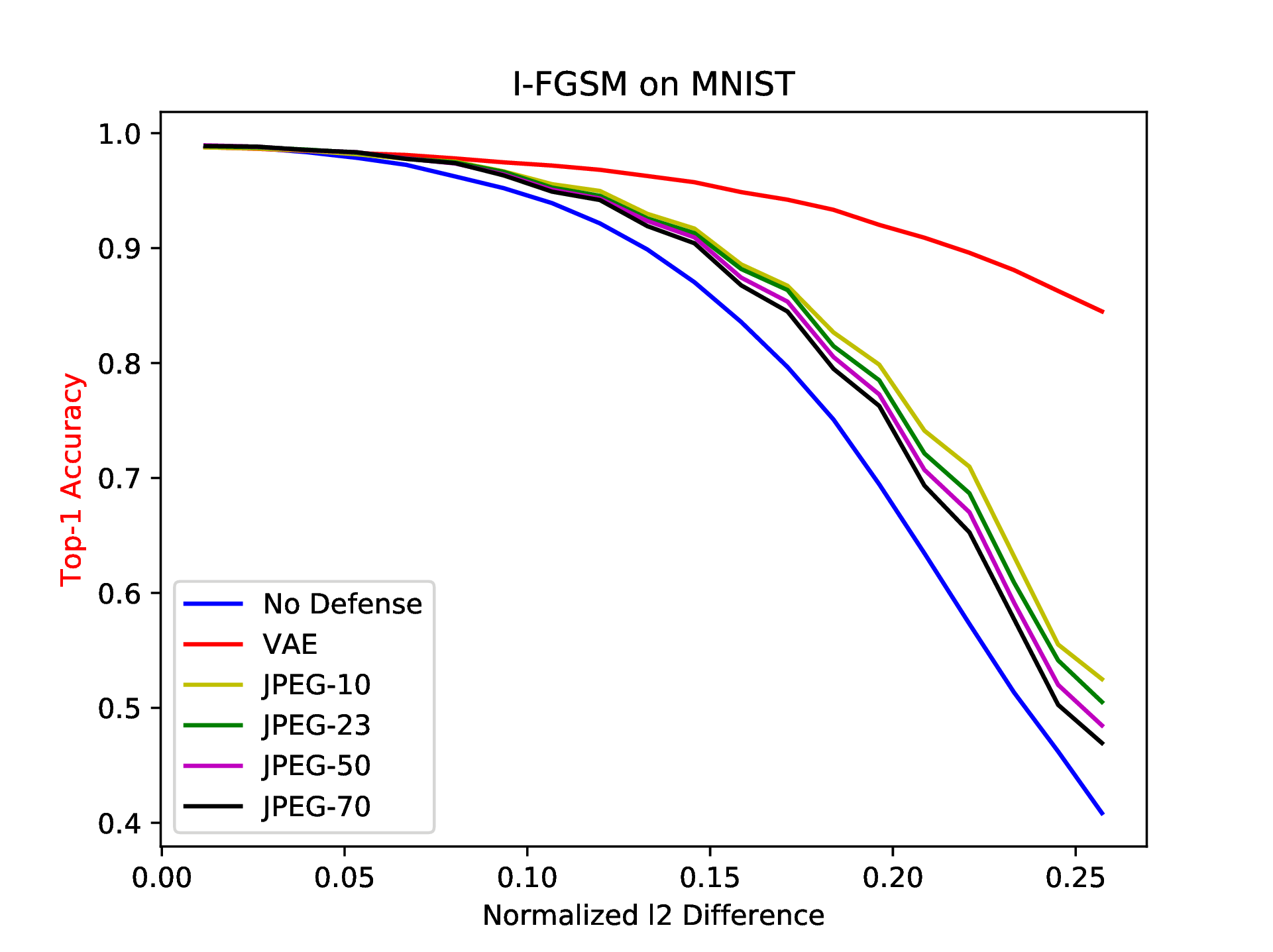} \\[\abovecaptionskip]
    \end{tabular}
    \caption{FGSM and I-FGSM attack on MNIST. Our VAE defense significantly outperforms JPEG compression of different qualities.}
    \label{fig:MNIST}
\end{figure*}

\begin{figure*}[!htbp]
    \centering
    \begin{tabular}{@{}cc@{}}
    \includegraphics[width=0.4\linewidth]{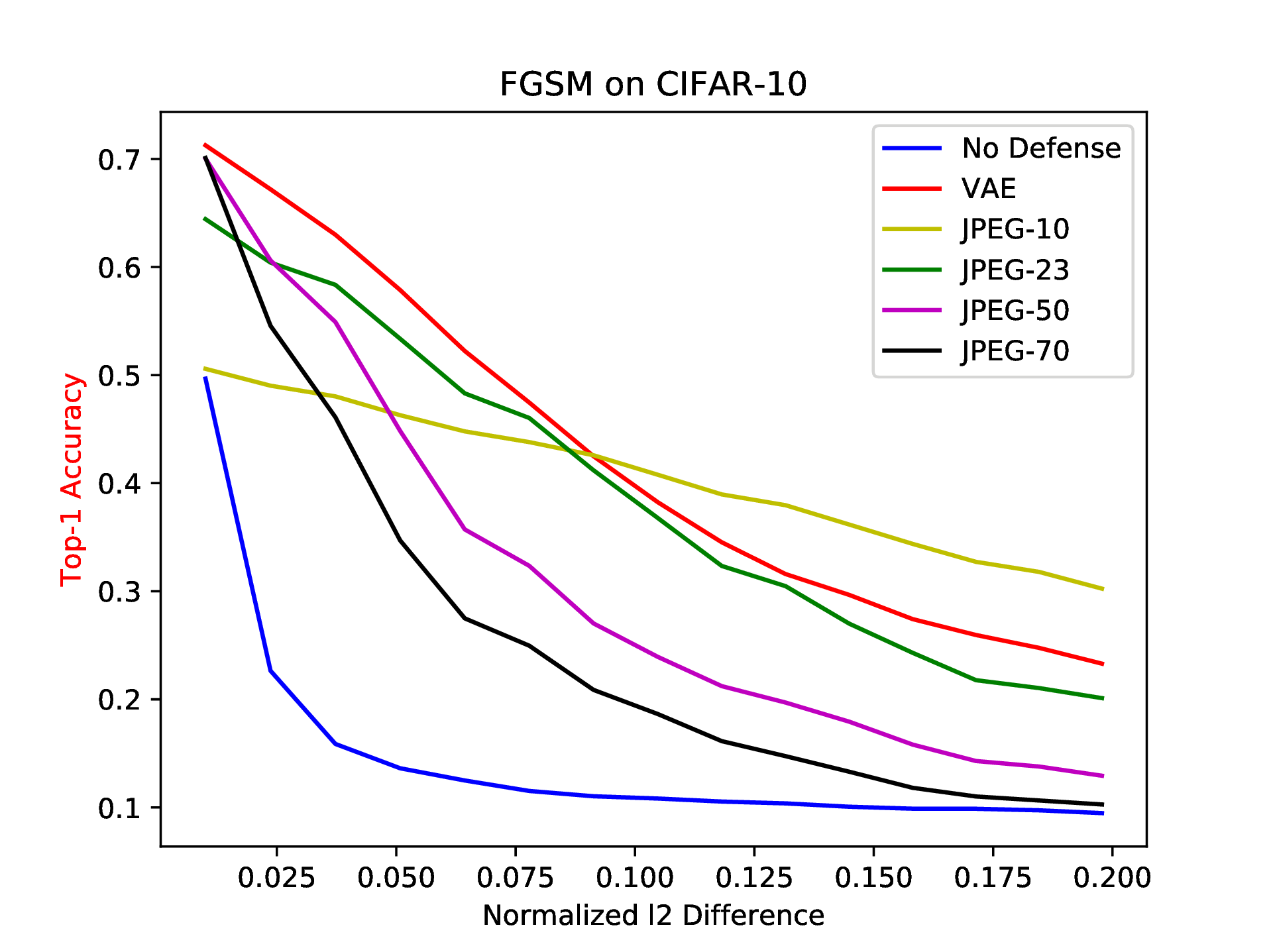} & \includegraphics[width=0.4\linewidth]{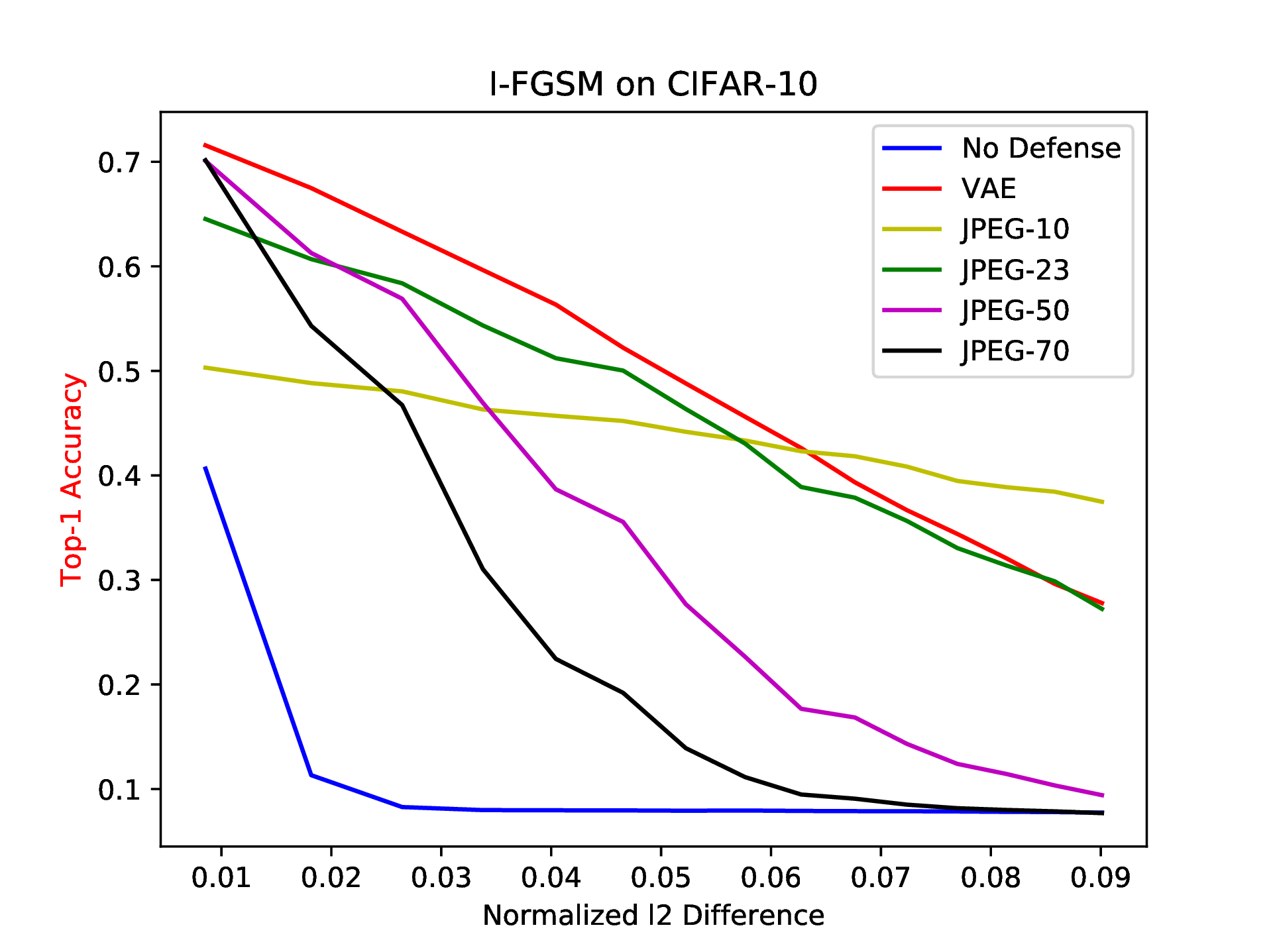} \\[\abovecaptionskip]
    \end{tabular}
    \caption{FGSM and I-FGSM attack on CIFAR-10. Our VAE defense outperforms JPEG compression in most cases. But, for low quality (10) and large perturbations, JPEG outperforms our VAE due to larger information loss, which also removes adversarial perturbations.}
    \vspace{-2mm}
\end{figure*}

\section{Experimental Setup}
In this section, we describe the datasets, CNN models, variational auto-encoder models, attacks and evaluation metrics used for our experiments.

\textbf{Datasets}. We use three different datasets: MNIST, CIFAR-10 and the NIPS 2017 Defense Against Adversarial Attacks development dataset\footnote{NIPS 2017 Defense Against Adversarial Attacks Kaggle Competition Dataset \hyperlink{1}{https://www.kaggle.com/c/nips-2017-defense-against-adversarial-attack/data}}. MNIST is a dataset of handwritten digits where each image is a $28 \times 28\times 1$ gray-scale image labeled one of ten classes from 0 to 9. There are 60000 training images and 10000 testing images. CIFAR-10 is a ten-class image dataset where each image is a $32 \times 32 \times 3$ RGB image. There are 50000 training images and 10000 testing images. The NIPS 2017 Defense Against Adversarial Attacks development dataset is a 1001-class image dataset where each image is $299 \times 299\ times 3$ RGB image obtained from Imagenet. We have access to the 1000 images provided by the competition in the development set.

\textbf{CNN Classifiers}. For MNIST and CIFAR-10, we set up our own CNN classifiers and train them for the classification. For NIPS dataset, we use the pre-trained 1001-class Inception V3 model \cite{szegedy2016rethinking} provided by the NIPS Kaggle competition\footnotemark[1]. The structure of each classifiers is described in Table \ref{tab:CNN}. Before being attacked by any adversarial attack, the classification accuracy on the MNIST test set, CIFAR-10 test set and NIPS 2017 dataset are 0.9904, 0.8663 and 0.945 respectively. We use these as baselines when evaluating our defense.

\textbf{Attacks}. All our attacks are white-box attacks. Thus, We assume that the attacker has access to the parameters of the deployed classifier, but is not aware of the defense strategies. In order to make our defense universal, we only train our VAEs to reconstruct original images instead of learning to reconstruct original images from attacked images (adversarial training). We use FGSM, I-FGSM as our attack methods. For MNIST, we set $\epsilon \in [0,0.12]$ for FGSM and I-FGSM, the number of iterations for I-FGSM is 10. For CIFAR-10, we set $\epsilon \in [0,0.1]$ for FGSM and I-FGSM, the number of iterations for I-FGSM is 10. For NIPS, we set $\epsilon \in [0.005,0.09]$ for FGSM, following the same settings as Shaham \etal \cite{shaham2018defending}.

\begin{figure*}[t]
\centering
    \includegraphics[width=0.85\linewidth]{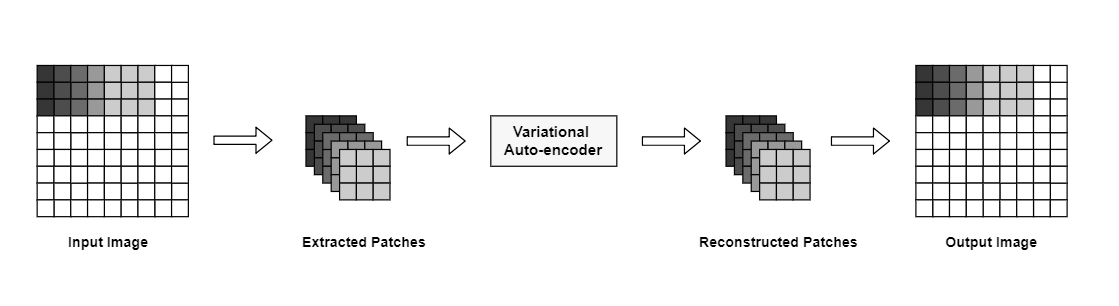}
    \vspace{-3mm}
    \caption{An schematic diagram of patch-wise reconstruction with $3 \times 3$ patches and a stride of 1.}
\label{fig:reconstruction}
\vspace{-2mm}
\end{figure*}

\textbf{Defense}. We evaluate our defense along with JPEG compression, which is the overall best basis transformation used as defense proposed by Shaham \etal\cite{shaham2018defending}. Following their results on tuning the JPEG quality parameter, we consider the JPEG quality of 23 and also evaluate the performance of other JPEG qualities.

\textbf{Evaluation}. We evaluate the performance of the attacks and defense on top-1 prediction accuracy versus the average relative L2 difference between original and attacked images. The average relative L2 difference is defined to be
$$
\text{L2 Diff} = \frac{1}{N}\sum_{i=1}^N\frac{||x_i - \hat{x}_i||_2}{||x_i||_2}.
$$
Normally a larger L2 difference results in larger perturbations which cause more errors in classification. However, for a more efficient attack, a relatively small L2 difference can result in a significant decrease in the classification accuracy. For example, in tables \ref{tab:NIPS_FGSM} and \ref{tab:NIPS_IFGSM}, a normalized relative L2 difference of 0.01 results in 0.35 and 0.15 accuracy for FGSM attack and I-FGSM attack respectively. I-FGSM is a more precisely crafted attack adding small perturbations that confuses the classifier more efficiently.

\begin{table*}[t]
\centering
\begin{tabular}{|cc|cc|c|c|cc|c|}
\hline
        &        & \multicolumn{2}{c|}{JPEG}       & VAE $16 \times 16$ p& VAE $32 \times 32$ p & \multicolumn{2}{c|}{VAE $64 \times 64$ p} &                \\
L2 Diff & No Def & JPEG 10        & JPEG 23        & Stride 16                & Stride 32                & Stride 32             & Stride 64             & Ensembled      \\ \hline
0.010   & 0.350  & 0.768          & \textbf{0.860} & 0.802                    & 0.782                    & 0.782                 & 0.762                 & 0.845          \\
0.023   & 0.235  & 0.743          & \textbf{0.792} & 0.764                    & 0.759                    & 0.767                 & 0.75                  & 0.789          \\
0.037   & 0.220  & 0.730          & 0.721          & 0.719                    & 0.724                    & 0.736                 & 0.731                 & \textbf{0.745} \\
0.052   & 0.213  & 0.709          & 0.652          & 0.675                    & 0.698                    & 0.709                 & \textbf{0.711}        & 0.689          \\
0.066   & 0.215  & 0.687          & 0.596          & 0.627                    & 0.677                    & 0.686                 & \textbf{0.692}        & 0.658          \\
0.095   & 0.232  & \textbf{0.652} & 0.501          & 0.576                    & 0.624                    & 0.647                 & 0.649                 & 0.59           \\
0.109   & 0.233  & 0.611          & 0.469          & 0.544                    & 0.592                    & \textbf{0.625}        & 0.622                 & 0.557          \\
0.124   & 0.240  & 0.602          & 0.461          & 0.532                    & 0.569                    & 0.61                  & \textbf{0.611}        & 0.537          \\
0.138   & 0.241  & 0.578          & 0.433          & 0.522                    & 0.569                    & 0.585                 & \textbf{0.588}        & 0.528          \\
0.152   & 0.247  & 0.569          & 0.426          & 0.497                    & 0.524                    & 0.564                 & \textbf{0.569}        & 0.509          \\
0.167   & 0.253  & 0.548          & 0.401          & 0.47                     & 0.512                    & 0.548                 & \textbf{0.563}        & 0.494          \\ \hline
\end{tabular}
\vspace{2mm}
\caption{Top-1 accuracies on NIPS 2017 dataset under FGSM attack.}
\label{tab:NIPS_FGSM}
\end{table*}

\section{Variational auto-encoder training}
\textbf{Models}. We trained multiple variational auto-encoder models from scratch. The encoder structures for MNIST and CIFAR-10 are in Table \ref{tab:VAE}. The decoder structure are mirror images of the encoders. For MNIST and CIFAR-10, we train the VAE on the training set images and evaluate the defense on the testing set.

For NIPS 2017 dataset, the size of original images are $299\times299\times3$. Instead of training an auto-encoder on the original images, we decided to train our VAE on patches randomly extracted from the 1000 original images provided. In this way, the model is easier to train and a patch-wise reconstruction is much more flexible. We use three different input patch sizes, $16\times16$, $32\times32$ and $64\times64$.

\textbf{Training}. We use the ADAM optimizer \cite{kingma2014adam} to perform the end-to-end training process independently of the attack. For the NIPS 2017 dataset, we also made some modifications to the model structure. In particular, a tunable clipping value to the noise $\epsilon \sim N(0,1)$ was added for sampling the latent space.
This restricts the random Gaussian noise to be within the range of clipping. We set the default value for clipping to [-5, 5] and used it for training. For each training process, we trained enough epochs until the first term in the loss function changes very little.

\section{Results}
\textbf{MNIST}. 
For MNIST, our defense is much more effective against the FGSM than JPEG. The results are consistent as we tune the FGSM and I-FGSM parameter to enhance the attack and cause a different amount of perturbations. In Table \ref{tab:MNIST}, for example, our VAE defense maintains an accuracy of 0.845 while JPEG (quality 23) defense can only defend the network from 0.408 to 0.505. The results for the I-FGSM attack are shown in Figure \ref{fig:MNIST}.

\begin{table*}[htbp]
\centering
\begin{tabular}{|cc|cc|c|c|cc|c|}
\hline
         &        & \multicolumn{2}{c|}{JPEG}       & VAE $16 \times 16$ p & VAE $32 \times 32$ p & \multicolumn{2}{c|}{VAE $64 \times 64$ p} &                \\
L2 Diff  & No Def & JPEG 10        & JPEG 23        & Stride 16                & Stride 32                & Stride 32             & Stride 64             & Ensembled      \\ \hline
0.01043  & 0.15   & 0.761          & \textbf{0.85}  & 0.808                    & 0.774                    & 0.781                 & 0.763                 & 0.846          \\
0.017752 & 0.05   & 0.751          & \textbf{0.804} & 0.772                    & 0.759                    & 0.764                 & 0.75                  & 0.803          \\
0.019472 & 0.044  & 0.752          & \textbf{0.806} & 0.783                    & 0.771                    & 0.76                  & 0.753                 & 0.796          \\
0.027197 & 0.033  & 0.735          & 0.769          & 0.758                    & 0.741                    & 0.755                 & 0.743                 & \textbf{0.778} \\
0.028969 & 0.032  & 0.728          & 0.763          & 0.747                    & 0.74                     & 0.757                 & 0.752                 & \textbf{0.776} \\
0.042603 & 0.025  & 0.719          & 0.706          & 0.701                    & 0.715                    & \textbf{0.739}        & 0.723                 & 0.736          \\
0.050261 & 0.024  & 0.712          & 0.657          & 0.675                    & 0.692                    & \textbf{0.723}        & 0.711                 & 0.689          \\
0.052266 & 0.024  & 0.703          & 0.651          & 0.673                    & 0.698                    & \textbf{0.714}        & 0.709                 & 0.699          \\
0.06017  & 0.022  & \textbf{0.704} & 0.611          & 0.629                    & 0.674                    & 0.698                 & 0.697                 & 0.653          \\
0.072137 & 0.022  & 0.678          & 0.526          & 0.609                    & 0.655                    & 0.676                 & \textbf{0.678}        & 0.619          \\
0.074035 & 0.02   & 0.678          & 0.526          & 0.614                    & 0.655                    & \textbf{0.679}        & 0.678                 & 0.623          \\ \hline
\end{tabular}
\vspace{2mm}
\caption{Top-1 accuracies on NIPS 2017 dataset under I-FGSM attack.}
\label{tab:NIPS_IFGSM}
\end{table*}

\textbf{CIFAR-10}.
For CIFAR-10, our defense also outperforms JPEG compression consistently in general. For small and large perturbations, our VAE defense maintains about $3\% \sim 6\%$ higher accuracy than JPEG with quality parameter 23. However, if we set the quality parameter to be comparatively low (10), at small adversarial perturbations JPEG compression restores a lot fewer images for correct classification due to the information loss. But, as the adversarial perturbation gets larger, JPEG surpasses the performance of VAE because the significant information loss also removes most of the adversarial perturbations.

\textbf{NIPS 2017}. For the NIPS 2017 dataset, we trained three different VAE models with different input image sizes, $16\times16$, $32\times32$ and $64\times64$. The training data are randomly extracted patches from the 1000 images. Since the original images are $299\times299\times3$, the training process would be very time consuming if we directly trained a variational auto-encoder on the images. In order to decrease training time and add flexibility, we decided to apply our image reconstruction on patches. Our approach is to extract patches from the original image, reconstruct the image with our trained VAE, stack reconstructed patches together and average the overlapping parts. 

Here, we introduce another hyper-parameter for our VAE defense: the stride of the reconstruction process where we stack overlapping pixels and take the average. A smaller stride results in more overlapping area and creates higher-quality images. But, it tends to also preserve the  adversarial perturbations. We also improve the defense's flexibility by adding this as a hyperparameter. Figure~\ref{fig:reconstruction} shows an example of image reconstruction with $3 \!\times\! 3$ patches and a stride of 1.

Reconstruction of images from patches typically suffers from sharp edges between patches stitched together. If we only stack overlapping patches and average the pixel values, we still suffer from significant artifacts near the corners and edges of the stacked patches. The worst effect is that these edge effects behave like additive noise for the classifier. To mitigate this problem, we also apply a $5 \times 5$ smoothing filter to every image we reconstruct from patches.  While the edge effects are significantly reduced, the smoothing filter also blurs the image. To compensate for this blurriness, we adjust the hyper-parameter $\beta$ and train the model longer to achieve more precise reconstruction of the images. The end result was that the smoothing filter increased the classification accuracy on VAE reconstructed images by roughly 10\%, for both benign and attacked images.

Table \ref{tab:NIPS_FGSM} shows the performance of our VAE defense on the NIPS 2017 dataset under FGSM attack. We apply VAE models of different input sizes with different reconstruction strides. Our best single model is the $64\times64$ patch VAE model. With an 0.167 average relative L2 difference between original and adversarial images, the model restores the classification accuracy back to 0.563. In this case, JPEG compression of quality 10 reaches an accuracy of 0.548. However, for small perturbations (0.01 L2 difference), JPEG compression of quality 23 significantly outperforms any other model, restoring the accuracy to 0.86. But the performance of 23 quality JPEG drops rapidly as the L2 difference increases.
Eventually, it ends up with an accuracy of 0.0401, which is the lowest. 

Through these experiments, we can observe that a better reconstruction of the original image results in better defense against small perturbations but fails to remove adversarial noise as the perturbations get larger. Our $16\times16$ and $32\times32$ patch VAE models show similar results and their defense against small perturbations is better than $64\times64$ patch model.
But, this advantage does not persist as the attack becomes severe. 

To overcome this disadvantages of each patch size, we also considered a ensemble-averaged output of the four VAE models in the table.  To compute this, the pixels of the four independent VAE reconstructions are averaged.  The last column shows the performance of the ensemble averaged output. Its performance approaches that of JPEG quality 10 on small perturbations while maintaining a better performance for larger perturbations.

The I-FGSM attack is a more efficient attack than the FGSM attack.
I-FGSM creates small perturbations (0.07 average relative L2 difference) that significantly decrease the classification accuracy to 0.02. As shown in Table \ref{tab:NIPS_IFGSM}, a single $64\times64$ patch VAE model is capable of restoring the classification accuracy from 0.025 to 0.739. From the results, we see that the ensemble-averaged model provides better defense against small perturbations than any single VAE model but sacrifices performance when facing larger perturbations.

\section{Conclusion}
In this paper, we explore the performance of variational auto-encoders (VAE) designed to defend against adversarial attacks and proposed a patch-wise reconstruction method for large resolution images. The proposed method turned out to be robust against FGSM and I-FGSM adversarial attacks.

In future work, we plan to improve these strategies by modifying the training to more directly reward the removal of adversarial attacks.  During our experiments, we observed that our method is less effective for small perturbations to large images. Thus, we are currently considering modified VAE structures to overcome this weakness.

{\small
\bibliographystyle{ieee}
\bibliography{references.bib}
}

\end{document}